\pdfoutput=1

\documentclass[11pt]{article}

\usepackage[preprint]{acl}

\usepackage{times}
\usepackage{latexsym}

\usepackage[T1]{fontenc}

\usepackage[utf8]{inputenc}

\usepackage{microtype}

\usepackage{inconsolata}

\usepackage{graphicx}

\title{\moelorafamily Methods for Multilingual Adaptation: A Systematic Empirical Study}

\newcommand{\moelorafamily}{MoE--LoRA\xspace}

\newcommand{\RQ}[1]{RQ\ensuremath{_{#1}}}

\newcommand{\eg}{e.\,g.,~}

\newcommand{\artifacturl}{https://github.com/dice-group/moe-study/tree/master}

\newcommand{\artifactlink}{\expandafter\url\expandafter{\artifacturl}}
\newcommand{\repourl}{https://github.com/dice-group/moe-study}
\newcommand{\repolink}{\expandafter\url\expandafter{\repourl}}
\newcommand{\artifactfile}[1]{\href{\artifacturl/#1}{\nolinkurl{#1}}}
\newcommand{\artifactnb}[1]{%
  \href{\artifacturl/paper_artifact/notebooks/#1}{\nolinkurl{#1}}}

\newcommand{\baseline}[1]{\textbf{#1}}

\newcommand{\gainmark}{\textcolor{teal}{$\blacktriangle$}}
\newcommand{\tiemark}{\textcolor{gray}{$\bullet$}}
\newcommand{\lossmark}{\textcolor{red!65!black}{$\blacktriangledown$}}

\author{
\bf Nikit Srivastava, Joel Dag, René Speck, Axel-Cyrille Ngonga Ngomo
\\ \\ Data Science Group, Heinz Nixdorf Institute, Paderborn University, Germany
\\ \\ \{nikit.srivastava, rene.speck, axel.ngonga\}@upb.de, joeldag@mail.upb.de \\
}

\usepackage{xspace}
\usepackage{float}
\usepackage{amsmath}
\usepackage{amssymb}
\usepackage{cleveref}
\usepackage{hyperref}
\usepackage{color}
\usepackage{multirow}
\usepackage{tabularx}
\usepackage{booktabs}
\usepackage{todonotes}
\usepackage{csquotes}

\usepackage[inline]{enumitem}

\usepackage[table]{xcolor}

\begin{document}
\maketitle

\begin{abstract}
Routed low-rank adapters offer a parameter-efficient means to improve the performance of large language models (LLMs) across tasks. Hence, it is assumed that they can also be used to improve the support of low-resource languages in LLMs, which currently serve English far better than the thousands of other languages spoken worldwide. However, this assumption has never been verified empirically under a shared multilingual evaluation protocol. We address exactly this gap by comparing eight low-rank adaptation (LoRA) methods on a frozen base model, using 146 training runs across 200 unique languages and more than 10{,}000 evaluations. Our results suggest that a reduced-scale search can identify viable configurations of LoRA methods but does not reliably predict their relative performance, as the top-ranking configuration is never best at scale. LoRA reliably improves language-model fit on translation data, consistently degrades reading comprehension, and benefits low-resource languages most. These findings suggest that LoRA methods should be regarded as targeted tools rather than general-purpose means for improving the multilingual capabilities of LLMs.
\end{abstract}

\section{Introduction}

Large language models (LLMs) are trained on data that are highly unevenly distributed across languages. English constitutes a substantial share of the training corpora of widely deployed models such as Llama 3, whereas many other languages are represented by far less data~\cite{llama3}. This disparity contributes to lower and less reliable performance beyond English, particularly for low-resource languages~\cite{ustun-etal-2024-aya,asai-etal-2024-buffet,ojo-etal-2025-afrobench}. Addressing this gap requires compute-efficient methods for extending existing models to many languages, rather than training a separate model for each one.

Continued pretraining (CPT) on multilingual data can add language coverage without training a new model from scratch. However, joint training also creates competition for limited model capacity. Updates that help one language can have little effect on another or can reduce its performance, a pattern described as \textit{negative interference} or the \textit{curse of multilinguality}~\cite{aharoni-etal-2019-massively,wu-dredze-2020-languages,wang-etal-2020-negative,conneau-etal-2020-unsupervised}. Multilingual adaptation therefore needs to balance parameter sharing, language-specific capacity, and training cost.

Running CPT~\cite{gururangan-etal-2020-dont,ke2023continual} at scale is expensive, which has made parameter-efficient fine-tuning (PEFT) the practical way to adapt large models. Low-Rank Adaptation (LoRA) freezes the base model and learns a single low-rank update~\cite{hu2022lora}. Because every language passes through those same parameters, LoRA cannot allocate capacity per language and leaves the interference problem largely unresolved. Mixture-of-Experts (MoE) extensions of LoRA target this limitation directly. They replace the single update with several low-rank experts that are shared or selected by a routing rule~\cite{10.1145/3626772.3657722,zadouri2024pushing}, adding capacity at low cost and letting different experts specialize. Such specialization can align with language structure. Indeed, routing in multilingual MoE models has been observed to correlate with implicit phylogenetic language groups~\cite{srivastava-etal-2025-lola}, making \moelorafamily a natural candidate for balancing shared and language-specific capacity during multilingual CPT. However, whether this additional structure actually leads to significantly better results remains an unsolved problem.

This question is all the more relevant when one considers that most \moelorafamily methods were developed and validated for multi-task settings such as medical adaptation and instruction tuning~\cite{10.1145/3626772.3657722,zadouri2024pushing}, rather than for multilingual learning. The few efforts that extend them to multilingual settings
each commit to a single design. MLAS-LoRA uses language-general and language-specific LoRA modules for machine translation~\cite{dong-etal-2025-mlas}, and MoE-LPR adds experts to a pretrained model with language-aware routing~\cite{10.1609/aaai.v39i24.34805}.
However, no prior work systematically compares \moelorafamily methods using the same base model, training data, and evaluation protocol for multilingual CPT. We address this gap through a controlled comparison that evaluates both aggregate performance and variation across tasks and languages.

Concretely, we compare eight parameter-efficient adaptation methods using a frozen Llama-3.1-8B base model. The search stages comprise 132 training runs on ten languages, including the longer follow-up on 22 combined configurations. We then carry the five best configurations, together with two plain-LoRA baselines, to fourteen broader runs covering 96 and 200 training languages, for 146 independent training runs in total. Four multilingual benchmarks evaluate reading comprehension, language-model fit on translation data, topic classification, and text classification, which gives each trained configuration around 70 benchmark and language measurements, 10{,}842 altogether.
\paragraph{Research questions.}
We organize the study around four questions:
\begin{enumerate}[leftmargin=1.5em,label=\textit{\arabic*.},labelsep=0.5em,itemsep=1pt,parsep=0pt,topsep=3pt]
\item \textit{How do rank and expert count affect multilingual performance?}
\item \textit{How do \moelorafamily methods compare under a shared multilingual evaluation?}
\item \textit{Do the strongest methods remain effective under broader multilingual training conditions?}
\item \textit{Which tasks and languages benefit most from \moelorafamily adaptation?}
\end{enumerate}

\paragraph{Contributions.}
We contribute a controlled comparison of routed low-rank experts under massively multilingual continued pretraining, which yields three key findings:
\begin{enumerate*}[label=(\roman*)]
\item low ranks win consistently, so the best-performing setting also trains the fewest parameters;
\item a reduced-scale search finds configurations that remain viable at scale but does not reliably predict how they rank there; and
\item what transfers instead is the direction of the effect, with gains concentrated on translation fit and on lower-resource languages.
\end{enumerate*}
We release the protocol, per-language results, and code so that a new method can enter the same comparison.\footnote{\repolink}
\section{Related Work}
In this section, abbreviations in bold indicate the baselines used in our experiments.
\paragraph{Mixture-of-Experts.} \citet{6797059} introduced a modular supervised learning framework in which multiple expert networks are trained on disjoint subsets of the data.
Following the transformer architecture~\cite{NIPS2017_3f5ee243}, sparse MoE became a primary approach for reducing the computational cost of LLMs~\citep{shazeer2017,JMLR:v23:21-0998,krajewski2024scaling}, and MoEs empirically outperform dense counterparts at a fixed computational budget, particularly at large scale and with well-optimized routing~\citep{zoph2022,NEURIPS2024_665bb142,10937907}.

\paragraph{Low-Rank Adaptation.} PEFT methods~\cite{ding2023parameter} attain comparable performance to full fine‑tuning while updating only a fraction of the parameters.
The most widely adopted of them, \baseline{LoRA}~\cite{hu2022lora}, trains only low‑rank perturbations to selected weight matrices, which reduces memory usage. It also generalizes robustly~\cite{xin-etal-2024-beyond} and better maintains the base model's performance outside the target domain~\cite{biderman2024lora}.
CoLA~\cite{zhou-etal-2025-cola} extends it with an efficient initialization scheme and three collaborative strategies between the low-rank matrices.

\paragraph{\moelorafamily Family.}
LoRA scales poorly to multi-task settings, where each task typically requires a separate adapter. To address this, \citet{10.1145/3626772.3657722} proposed \baseline{MOELoRA} for multi-task medical applications, and numerous \moelorafamily methods have followed with different routing strategies and expert architectures.

Initial contributions proposed dynamic LoRA composition and highly parameter‑efficient MoE for instruction tuning~\cite{huang2024lorahub,feng-etal-2024-mixture,zadouri2024pushing}, followed by token‑level and layer‑wise variants~\cite{buehler2024x}, multi‑task ones, first proposed for vision with MTLoRA~\cite{agiza2024mtlora}, and routing designed to mitigate world‑knowledge forgetting~\cite{dou-etal-2024-loramoe,wu2024mixture}.
Among the methods we compare, \baseline{AdaMoLE}~\cite{liu2024adamole} and \baseline{AdaMoE}~\cite{zeng-etal-2024-adamoe}\footnote{\baseline{AdaMoE} implementation builds on the MoLA formulation~\cite{gao-etal-2025-mola}, so the released configurations and run identifiers carry a \texttt{mola} prefix.} select experts adaptively through threshold networks or null‑expert designs, \baseline{HydraLoRA}~\cite{tian2024hydralora} reduces redundancy through an asymmetric structure, \baseline{HMoRA}~\cite{liao2025hmora} combines token‑ and task‑level routing across layers, \baseline{MTL-LoRA}~\cite{yang25} separates task‑specific from shared representations, and \baseline{MoV}~\cite{zadouri2024pushing} mixes vectors that inhibit and amplify inner activations~\cite{NEURIPS2022_0cde695b}.

These methods target instruction tuning and multi-task learning, not languages. Multilingual adaptation has instead relied on modular methods such as MAD-X~\cite{pfeiffer-etal-2020-mad}, X-MOD~\cite{pfeiffer-etal-2022-lifting}, and MLAS-LoRA~\cite{dong-etal-2025-mlas}, which assign modules per language by construction rather than by routing. Routed multilingual work mostly targets speech recognition~\cite{song24_interspeech,bagat25_interspeech,11030302}, and MoE-LPR routes between full feed-forward rather than low-rank experts~\cite{10.1609/aaai.v39i24.34805}. LoRALib benchmarks \moelorafamily methods, but not for multilingual adaptation~\cite{wang2025loralib}. Routed low-rank experts under a shared multilingual CPT protocol at several language scales remain unstudied.

\paragraph{Multilingual Language Modeling.}
Early multilingual models showed that a single pretrained transformer can learn shared representations across dozens of languages~\cite{devlin-etal-2019-bert,conneau-etal-2020-unsupervised}, and large-scale multilingual LLMs have since extended this to hundreds of languages~\cite{JMLR:v25:23-0581,srivastava-etal-2025-lola,yang2025qwen3technicalreport,gemmateam2026gemma4technicalreport}.

A complementary line of work uses CPT as a cheaper route to multilinguality. Sailor2~\cite{dou2025sailor2sailingsoutheastasia} applies block expansion and CPT on roughly 500 billion tokens for 15 Southeast Asian languages, and comparable efforts cover Japanese, Korean, and Indic languages~\cite{fujii2024continual,vo2024redwhaleadaptedkoreanllm,nag-etal-2025-efficient}. Others study how language mixing interacts with CPT~\cite{li2025rethinking} or target selected layers with LoRA when adding new languages~\cite{owodunni2026continually}. We apply solutions from the multi-task domain to this setting and compare \moelorafamily methods across languages and tasks.

\section{Methodology}\label{sec:method}
\subsection{Problem Setting}
Let $\Phi$ denote the parameters of a pretrained autoregressive language model and $\mathcal{D}$ a CPT corpus of token sequences drawn from $K$ languages. The goal is to adapt the model to the joint distribution of all $K$ languages while mitigating catastrophic forgetting~\cite{doi:10.1073/pnas.1611835114} of the capabilities it already has. Rather than updating $\Phi$, we keep it frozen and train a separate set of parameters $\Theta$ on top of it (Section~\ref{method:trainable_params}).

\subsection{Causal Language Model Objective}
Given a sequence $s = (x_1,x_2,\dots,x_{|s|})$ of $|s|$ tokens, the CLM objective maximizes the probability of each token conditioned on its causal prefix $x_{<t} = (x_1,\dots,x_{t-1})$ using the standard autoregressive next-token prediction mechanism and cross-entropy loss over target words~\cite{bengio2003, NIPS2017_3f5ee243}:
\[
    \mathcal{L}_{\text{CLM}}(\Theta)
    = -\sum_{s\in\mathcal{D}}\; \sum_{t=1}^{|s|}
      \log P_\Theta\! \bigl(x_t \mid x_{<t}\bigr),
\]
where $\Theta$ is the set of trainable parameters (Section~\ref{method:trainable_params}) and $P_\Theta$ is the usual softmax over the output projection of the final hidden state.%

\subsection{\moelorafamily Architectures}
\subsubsection{Expert Modules}
All pretrained weights $\Phi$ are frozen~\cite{dou-etal-2024-loramoe, 10.1609/aaai.v39i24.34805}. For each target weight matrix $W_0^{(\ell)}\in\mathbb{R}^{d_{\text{out}}\times d_{\text{in}}}$ at layer $\ell \in \{1,\dots,M\}$, we introduce $E$ low‑rank expert modules following the LoRA formulation~\cite{hu2022lora}:
\[
    \mathcal{E}_e^{(\ell)}(x) = B_e^{(\ell)} A_e^{(\ell)} x,
\]
where for each expert $e\in\{1,\dots,E\}$, $A_e^{(\ell)}\in\mathbb{R}^{r\times d_{\text{in}}}$ is the down‑projection, $B_e^{(\ell)}\in\mathbb{R}^{d_{\text{out}}\times r}$ the up‑projection, $r\ll\min(d_{\text{in}},d_{\text{out}})$  the rank, and $x\in\mathbb{R}^{d_{\text{in}}}$ the input to the layer.

\subsubsection{Routing and Aggregation}
The adapted layer output $h^{(\ell)}$ combines the frozen weights with a weighted sum of experts, utilizing the LoRA scaling factor $\alpha/r$~\cite{hu2022lora} and a sparse routing mechanism~\cite{shazeer2017,JMLR:v23:21-0998}:
\[
    h^{(\ell)} = W_0^{(\ell)} x + \frac{\alpha}{r} \sum_{e=1}^{E} \omega_e^{(\ell)}(x) \cdot \mathcal{E}_e^{(\ell)}(x),
\]
where $\alpha$ is the LoRA scaling hyperparameter, tied to the rank as described in Section~\ref{sec:setup-model}, and $\omega_e^{(\ell)}(x)\ge 0$ is the routing weight for expert $e$, determined by a routing function that may condition on the token representation $x$, a language identifier, the layer index $\ell$, or any combination thereof.

\subsubsection{Trainable Parameters} \label{method:trainable_params}
The set of trainable parameters is defined as:
\[
    \Theta \;=\; \bigcup_{\ell=1}^{M}
               \Bigl(
                 \{\,A_e^{(\ell)},\,B_e^{(\ell)}\}_{e=1}^{E}
                 \;\cup\;
                 \Theta_{\text{route}}^{(\ell)}
               \Bigr),
\]
where $\Theta_{\text{route}}^{(\ell)}$ contains the routing mechanism parameters (\eg gating weights, threshold networks, or language embeddings) at layer $\ell$, and $M$ is the total number of transformer layers.  Pretrained weights $\Phi$ receive no gradient.

\subsubsection{Training Objective}
To prevent expert collapse and ensure balanced utilization, the CLM loss is augmented with an auxiliary loss $\mathcal{L}_{\text{aux}}$ as described in sparse MoE literature~\cite{shazeer2017,JMLR:v23:21-0998}:
\[
    \mathcal{L}_{\text{total}}(\Theta) = \mathcal{L}_{\text{CLM}}(\Theta) + \lambda \cdot \mathcal{L}_{\text{aux}}(\Theta),
\]
where $\lambda$ is a balancing coefficient and $\mathcal{L}_{\text{aux}}$ is an auxiliary loss that regularizes the routing distribution (\eg encouraging balanced expert utilization or routing certainty).  Gradients of $\mathcal{L}_{\text{total}}$ flow through the expert modules and routing parameters only, with each expert's learning signal scaled by its routing weight $\omega_e^{(\ell)}(x)$.%
\section{Experimental Setup}

The study runs in three stages, described in Section~\ref{sec:stages}. The first two are exploratory, because they use the same observations both to select and to compare configurations.

The design is organized by the two quantities of Section~\ref{sec:method}, the per-expert rank $r$ and the expert count $E$. We use \emph{expert count} throughout for what some methods call heads or projections, and where a method divides a total rank budget among its experts.
We report the per-expert value, so that a total rank of 64 over 32 experts appears as $r=2$. Configurations are named by method, $r$, $E$, and top-$k$, the number of experts routed per token, so MOELoRA r2/16 k8 has $r{=}2$, $E{=}16$, and $k{=}8$.

\subsection{Model and Training Data}\label{sec:setup-model}
All experiments start from the frozen \texttt{meta-llama/Llama-3.1-8B} checkpoint~\cite{llama3}, which supplies the pretrained parameters $\Phi$ of Section~\ref{sec:method}.
English training text comes from FineWeb and the remaining languages from FineWeb2~\cite{NEURIPS2024_370df50c,penedo2025fineweb}. We use training sets of $K=10$, $96$, or $200$ languages (see Appendix~\ref{sec:appendix-data}). Each set takes the same fraction of every language's available documents, so languages keep their relative sizes in FineWeb2 and low-resource languages receive no extra weight. Table~\ref{tab:language-mixture} gives the resulting proportions.

Unless a method requires a different adapter layout, we attach adapters to the attention and feed-forward projections of every layer. The shared settings include a dropout rate of $0.05$, a learning rate of $2\!\times\!10^{-4}$, BF16 training, a sequence length of $1,024$, and a seed of $42$. We set $\alpha=2r$~\cite{biderman2024lora}, so the scaling $\alpha/r$ stays at $2$ across ranks for every method except HMoRA (Appendix~\ref{sec:appendix-budget}). The base model stays frozen, so training updates only the expert and routing parameters $\Theta$, and every run uses four NVIDIA A100 40GB GPUs. Settings that only one method defines, such as its routing rule, are recorded per run in the released configuration files.

\subsection{Methods}
The candidate pool covers AdaMoE, AdaMoLE, HMoRA, HydraLoRA, LoRA, MOELoRA, MTL-LoRA, and MoV, which routes vectors rather than low-rank matrices but demonstrates stronger performance than LoRA~\cite{zadouri2024pushing}. They represent different ways of sharing or routing experts, and each can be trained in a single multilingual run.

The search space is not a complete factorial grid, because not every method defines every factor and the number of settings per method differs. We therefore restrict controlled comparisons to configurations with shared coverage and give the full matrix in Appendix~\ref{sec:appendix-coverage}. Plain LoRA takes part in the shared rank panel, and its two best ranks there also serve as baselines in both broader regimes. They help separate the effect of the multi-expert structure from that of adaptation itself.

\subsection{Evaluation}\label{exp_set:evaldataset}
We evaluate the final checkpoint with the LM Evaluation Harness~\cite{lintang_sutawika_2026_18636344} on four multilingual benchmarks. Belebele (Bele.) measures reading comprehension~\cite{bandarkar-etal-2024-belebele}, FLORES (FLO.) supplies parallel translation data~\cite{nllbteam2022languageleftbehindscaling}, SIB-200 (SIB) measures topic classification~\cite{adelani-etal-2024-sib}, and Taxi1500 (Taxi) measures text classification~\cite{ma-etal-2025-taxi1500}. Together they cover both task accuracy and language-model fit.

We count a \emph{training run} as one configuration trained once and an \emph{evaluation run} as one model measured on one benchmark and language pair, counting the two FLORES directions separately (Table~\ref{tab:study-funnel}).

We use accuracy for Belebele, SIB-200, and Taxi1500. FLORES is evaluated as a language-modeling task rather than by decoding, so we score it with bits-per-byte over the reference translation~\cite{gao2020pile}. Normalizing by bytes rather than tokens keeps this score comparable across languages. It measures how well a model \emph{fits} translation data~\cite{NEURIPS2024_760b2d94}, so a FLORES gain is not evidence of better translation output. Lower values are better on FLORES, while higher values are better on the other three. Therefore, we invert FLORES before any comparison.

\subsection{Experimental Stages}\label{sec:stages}
\paragraph{One-factor search (\RQ{1}).}
We change one design choice at a time, first varying the rank $r$ while holding each method's expert count fixed and then varying the expert count $E$ while holding the rank fixed. The balanced rank analysis uses $r \in \{2, 4, 8, 16, 32, 64\}$ across the six methods that cover every one of those values, and the expert-count analysis uses $E \in \{2, 4, 8, 16, 32\}$.

\paragraph{Mixed follow-up (\RQ{2}).}
The first stage leaves the two best values of each factor, $r \in \{2, 8\}$ and $E \in \{16, 32\}$. Combining them per method and removing duplicates gives 22 configurations, which we train for 7.2 hours on the same ten-language data (Appendix~\ref{sec:appendix-rq2}). Eight are new rank--expert combinations and the remaining fourteen repeat an earlier sweep setting at the longer budget, which separates the effect of combining choices from the effect of simply training longer. The five best benchmark-balanced configurations go to the broader validation, so that the broader regimes test the shortlist itself rather than a single bet on the top position.

\paragraph{Broader validation (\RQ{3}--\RQ{4}).}
We train each of the five promoted configurations, plus plain LoRA at $r{=}2$ and $r{=}4$, once with 96 training languages (34.5 hours per run) and once with 200 training languages (72 hours per run), and evaluate them against the unadapted base model. Runs within a regime are matched on wall clock, hardware, data, and optimizer settings, which are the usual constraints in practice. Equal wall clock is not equal data exposure, because methods differ in throughput. Details of resource accounting are provided in Table~\ref{tab:resource-accounting}. The two regimes differ in several factors at once, so we compare systems, that is, the adapted configurations and the base model, \emph{within} each regime only.

\subsection{Aggregation and Statistical Tests}\label{sec:aggregation}
We reserve $r$ for the rank of a low-rank expert and use the term \emph{average position} for the ordering of systems, which runs from 1 for the best system downwards. We compare systems only on matching benchmark, task, language, and metric cases, and within each case we order the systems from best to worst and record the position of each. We then average the two FLORES directions and give each of the four benchmark families equal weight, which keeps benchmarks with more evaluation cases from dominating the comparison.

For \RQ{1} and \RQ{2} we use one Friedman test per benchmark and report Kendall's $W$, which measures how far the individual evaluation cases agree on an ordering, from 0 (they disagree entirely) to 1 (every case gives the same ordering), so that a low $W$ means a benchmark does not separate the systems rather than that the systems are equal. Follow-up pairwise comparisons use two-sided Wilcoxon signed-rank tests, with each stated family of tests adjusted using Holm's method. We call a result significant when its $p$-value, Holm-adjusted ($p_{\mathrm{Holm}}$) within such a family, is below $.05$.

In each broader regime we run one Friedman test per benchmark over all eight systems, followed by two-sided paired Wilcoxon tests on \emph{every} one of the twenty-eight system pairs, with Holm correction inside each regime and benchmark. A non-significant result means only that the available evaluation does not distinguish two systems, and never that those systems are equivalent.
\section{Results}
All comparisons are reported as average positions (Section~\ref{sec:aggregation}), where position 1 is best, so lower is always better, and the value says where a system places relative to the others rather than by how much. For brevity, the complete panels and the per-comparison statistics are in Appendix~\ref{sec:appendix}, as tables numbered A1 onwards, alongside all the related details.

\subsection{Rank and Expert Count (\RQ{1})}\label{sec:rq1}
We varied one factor at a time on the ten-language data, comparing on balanced panels, that is, restricted to the values at which all compared methods were actually run. Table~\ref{tab:rq1-full-ranks} gives the average positions and Table~\ref{tab:rq1-omnibus} the tests behind this section.

\paragraph{Low ranks win, and they are also the smallest.}
A rank of $r=2$ has the best overall average position (2.49), with $r=8$ second (2.72). Rank 2 leads on FLORES, SIB-200, and Taxi1500, whereas rank 8 leads on Belebele. Since adapter size grows with $r$, the winning setting is also the smallest. Rank 2 trains roughly an order of magnitude fewer adapter parameters than rank 64. How much rank matters depends strongly on the benchmark, as the evaluation cases agree closely on the rank ordering on FLORES ($W=0.63$) and barely at all on SIB-200 ($W=0.05$). The tests establish only that the ranks differ from each other, not that $r=2$ beats $r=8$.

\paragraph{Expert count has a weaker and flatter effect.}
$E=32$ takes the best average position (2.59) ahead of $E=16$ (2.75). Agreement between evaluation cases stays between $W=0.10$ and $W=0.15$ on all four benchmarks, against a range of 0.05 to 0.63 for rank. Expert count therefore matters less than rank, and it matters about equally little on every benchmark.

\subsection{\moelorafamily Methods (\RQ{2})}\label{sec:rq2}
\begin{table}[t]
\centering
\small
\setlength{\tabcolsep}{5pt}
\begin{tabular}{lrrrrr}
\toprule
Method & Bele. & FLO. & SIB & Taxi & Overall \\
\midrule
MOELoRA & \textbf{2.04} & 3.32 & 2.81 & \textbf{2.56} & \textbf{2.68} \\
AdaMoE & 2.62 & \textbf{2.80} & \textbf{2.63} & 2.83 & 2.72 \\
HydraLoRA & 3.36 & 3.12 & 2.74 & 3.79 & 3.25 \\
MTL-LoRA & 3.78 & 2.97 & 4.18 & 3.75 & 3.67 \\
LoRA & 3.35 & 2.88 & 4.76 & 3.93 & 3.73 \\
HMoRA & 5.85 & 5.91 & 3.88 & 4.14 & 4.94 \\
\bottomrule
\end{tabular}
\caption{Average position of the six methods with complete coverage of ranks 2--64 on Belebele, FLORES, SIB-200 and Taxi1500. Lower is better, best per column in bold. Coverage matrix in Table~\ref{tab:config-grid}.}
\label{tab:rq2-family}
\end{table}

\begin{table*}[t]
\centering
\small
\setlength{\tabcolsep}{10pt}
\begin{tabular}{lr@{\,}lr@{\,}lr@{\,}lr@{\,}lc}
\toprule
System & \multicolumn{2}{c}{Belebele} & \multicolumn{2}{c}{FLORES} & \multicolumn{2}{c}{SIB-200} & \multicolumn{2}{c}{Taxi1500} & Overall \\
\midrule
\multicolumn{10}{l}{\textit{96 training languages}} \\
MOELoRA r2/32 k8 & 4.25 & \lossmark & \textbf{3.35} & \gainmark & \textbf{1.49} & \gainmark & 5.36 & \tiemark & \textbf{3.61} \\
LoRA r2 & 4.08 & \lossmark & 4.38 & \gainmark & 4.19 & \tiemark & \textbf{2.86} & \gainmark & 3.88 \\
MOELoRA r2/16 k8 & 4.28 & \lossmark & 4.11 & \gainmark & 4.14 & \tiemark & 3.04 & \gainmark & 3.89 \\
MTL-LoRA r8/16 & 5.22 & \lossmark & 3.76 & \gainmark & 5.92 & \lossmark & 3.27 & \gainmark & 4.54 \\
Base model & \textbf{2.08} &  & 6.87 &  & 4.00 &  & 5.24 &  & 4.55 \\
HydraLoRA r2 & 4.00 & \lossmark & 5.28 & \gainmark & 3.59 & \tiemark & 6.12 & \tiemark & 4.75 \\
AdaMoE r2/8 k6 & 6.30 & \lossmark & 4.23 & \gainmark & 5.76 & \lossmark & 4.53 & \tiemark & 5.21 \\
LoRA r4 & 5.80 & \lossmark & 4.02 & \gainmark & 6.91 & \lossmark & 5.58 & \tiemark & 5.58 \\
\addlinespace
\multicolumn{10}{l}{\textit{200 training languages}} \\
HydraLoRA r2 & 4.35 & \lossmark & \textbf{2.62} & \gainmark & \textbf{1.64} & \gainmark & 5.18 & \tiemark & \textbf{3.45} \\
LoRA r4 & 5.71 & \lossmark & 4.28 & \gainmark & 5.36 & \lossmark & \textbf{1.41} & \gainmark & 4.19 \\
Base model & \textbf{1.79} &  & 6.04 &  & 3.54 &  & 5.59 &  & 4.24 \\
MOELoRA r2/32 k8 & 3.89 & \lossmark & 4.77 & \gainmark & 4.01 & \tiemark & 4.51 & \tiemark & 4.29 \\
MTL-LoRA r8/16 & 4.77 & \lossmark & 5.54 & \gainmark & 3.44 & \tiemark & 4.72 & \tiemark & 4.62 \\
LoRA r2 & 4.01 & \lossmark & 3.14 & \gainmark & 7.78 & \lossmark & 3.91 & \gainmark & 4.71 \\
AdaMoE r2/8 k6 & 6.47 & \lossmark & 3.48 & \gainmark & 3.88 & \tiemark & 5.98 & \tiemark & 4.95 \\
MOELoRA r2/16 k8 & 5.01 & \lossmark & 6.14 & \tiemark & 6.36 & \lossmark & 4.70 & \tiemark & 5.55 \\
\bottomrule
\end{tabular}
\caption{Each system within its training regime, showing its average position on every benchmark (lower is better, best per column in bold) followed by the outcome of its paired comparison against the base model, \gainmark~better, \tiemark~not distinguishable, \lossmark~worse. Underlying tests in Tables~\ref{tab:rq3-vs-baseline-tests} and~\ref{tab:rq3-omnibus}.}
\label{tab:rq3-final}
\end{table*}

\begin{table}[t]
\centering
\small
\begin{tabular}{lccc}
\toprule
Configuration & 10 & 96 & 200 \\
\midrule
MTL-LoRA r8/16 & 1 & 3 & 3 \\
MOELoRA r2/16 k8 & 2 & 2 & 5 \\
HydraLoRA r2 & 3 & 4 & 1 \\
MOELoRA r2/32 k8 & 4 & 1 & 2 \\
AdaMoE r2/8 k6 & 5 & 5 & 4 \\
\bottomrule
\end{tabular}
\caption{Position of each promoted configuration among the five, at the selection stage and in the two broader regimes. Columns give the number of training languages.}
\label{tab:rq3-transfer}
\end{table}

Two comparisons answer \RQ{2}: the first holds rank coverage fixed and compares methods (Table~\ref{tab:rq2-family}), while the second combines the values selected above and produces the shortlist from the 22 configurations of that stage (Appendix~\ref{sec:appendix-rq2}).

\paragraph{At matched ranks, MOELoRA and AdaMoE lead.}
Table~\ref{tab:rq2-family} compares the six methods that were run at every rank from 2 to 64, which is the largest set of methods we can compare without letting differences in grid coverage decide the outcome. MOELoRA and AdaMoE take the best overall average positions (2.68 and 2.72) ahead of HydraLoRA (3.25), and the per-benchmark leaders differ, with MOELoRA leading on Belebele and Taxi1500 and AdaMoE on FLORES and SIB-200. MoV has no rank to vary and AdaMoLE was not run in the rank search, so this panel compares six methods rather than summarizing the full search.

\paragraph{After combining factors, the shortlist is tight.}
The longer follow-up identifies five leading configurations within a narrow band of average positions among the 22 evaluated: MTL-LoRA (7.28), MOELoRA with 16 experts (7.48), HydraLoRA (7.50), MOELoRA with 32 experts (8.18), and AdaMoE (9.24). Their performance profiles vary by task, but only FLORES distinguishes the top three, with MOELoRA outperforming both MTL-LoRA ($p_{\mathrm{Holm}}=.011$) and HydraLoRA ($p_{\mathrm{Holm}}=.002$). The other three benchmarks show no significant differences among them. Given the narrow overall spread and limited task-level separation, we carry all five configurations forward rather than selecting only the top-ranked one.%

\subsection{Broader Multilingual Validation (\RQ{3})}\label{sec:rq3}
Table~\ref{tab:rq3-final} orders the five promoted configurations, the two plain-LoRA baselines, and the unadapted base model within each regime, marking each position with the outcome of that system's direct test against the base model, which we return to in \RQ{4}.

\paragraph{The unadapted model is difficult to outperform.}
The base model places fifth of eight after training on 96 languages and \emph{third} of eight after training on 200. No adapted configuration is best on more than two of the four benchmarks in either regime, and none is best in both regimes. Adaptation with these methods is therefore not a general improvement over the starting model, and performance varies between tasks, which we quantify in \RQ{4}.

\paragraph{Which method wins depends on the regime.}
HydraLoRA is fifth of the seven adapted configurations at 96 training languages and first at 200, where it is significantly better than the base model on FLORES and SIB-200. MOELoRA with 16 experts moves in the opposite direction, from third to last. MOELoRA with 32 experts leads at 96 languages. Since the regimes differ in several factors at once, we describe this as an interaction between method and regime rather than an effect of language count. Appendix~\ref{sec:appendix-rq3} reports the omnibus test behind each benchmark column and the pairwise comparisons against the base model.
\paragraph{The selection stage did not predict either outcome.}
Table~\ref{tab:rq3-transfer} follows each configuration across the three stages. The configuration ranked first after the ten-language follow-up finishes third in both broader regimes, and the two configurations that end at the extremes at 200 languages were adjacent in the middle of the shortlist. Spearman correlations between the stage orderings are $\rho=0.30$ (ten languages to 96), $\rho=-0.10$ (ten languages to 200), and $\rho=0.00$ (96 to 200). With five configurations these values are rough indicators, but they match the tables, which show little rank-order agreement across stages. The reduced-scale search identified a band of workable configurations without identifying the ordering inside it.

\subsection{Benefits and Regressions by Task and Language (\RQ{4})}\label{sec:rq4}
The marks in Table~\ref{tab:rq3-final} reduce the comparison against the base model to one verdict per configuration, regime, and benchmark. \RQ{1}--\RQ{3} ask only which system ranks highest, which average position answers. \RQ{4} asks how much adaptation gives and takes, on benchmarks whose units do not compare, so we add two measures to the same tests. One is the rank-biserial effect, the share of ranked evidence for a system minus the share against it, ranging from $-1$ to $+1$. The other highlights each difference against the base model's own level. Table~\ref{tab:rq3-vs-baseline-tests} lists the medians, effects, and adjusted $p$ values.

\begin{figure*}[t]
\centering
\includegraphics[width=\linewidth]{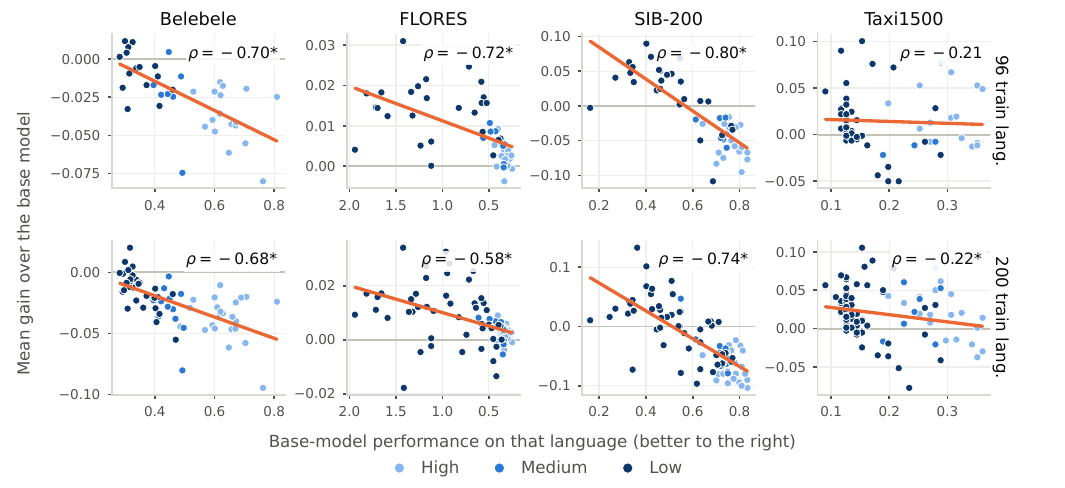}
\caption{Per-language gain over the base model against its own score on that language, averaged over the seven adapted configurations. Each point is one language, coloured by its resource level.}
\label{fig:rq4-gain-vs-base}
\end{figure*}

\paragraph{Adaptation is not uniformly beneficial.}
All fourteen configuration--regime combinations, plain LoRA included, are significantly \emph{worse} than the base model on Belebele, where the median regression runs from 1.4 to 3.4 accuracy points, roughly 7 to 18\% of the base model's margin above chance. Thirteen of fourteen are significantly \emph{better} on FLORES, where the median gain of 0.002 to 0.006 bits per byte is a relative reduction of 0.4 to 1.4\%. The effect sizes are what can be compared across the two scales: at 96 languages those on FLORES (+0.73 to +0.98) mirror the Belebele ones ($-0.78$ to $-0.95$), so the benefit is as consistent as the regression, while at 200 languages it attenuates and becomes configuration-dependent, holding for HydraLoRA, AdaMoE, and both plain-LoRA baselines (+0.67 to +0.83) while the other three shrink towards zero (+0.17 to +0.40). SIB-200 is where the methods actually diverge, with outcomes ranging from significantly better to significantly worse depending on configuration and regime. On Taxi1500 no configuration is significantly worse than the base model, but at 200 languages only the two plain-LoRA baselines are significantly better, and none of the five routed configurations can be distinguished from it. Because each configuration is trained once, these tests describe consistency across evaluation cases and not uncertainty across training runs.

\paragraph{The gain tracks how weak the base model already was.}

In Figure~\ref{fig:rq4-gain-vs-base} the languages the base model handles worst sit on the left, and vertical position is how much adaptation changes that score, so languages above the grey line improved and those below degraded. The orange line is the trend and $\rho$ the rank correlation behind it, marked * where $p<.05$.

That line slopes downward in all eight panels, with $\rho$ between $-0.58$ and $-0.80$ on Belebele, FLORES, and SIB-200 in both regimes, and only weakly on Taxi1500 ($-0.21$ and $-0.22$). Adaptation therefore redistributes capability rather than adding it uniformly, giving most to the languages the base model served worst and least, or nothing, to the languages it already served well. The three benchmarks share that trend and differ only in where it sits relative to the grey line, high enough on FLORES that nearly every language gains, crossing it on SIB-200, and below it on Belebele. The Taxi1500 slope is close to flat, which is consistent with its inability to separate systems elsewhere. Low-resource languages cluster on the left, so the coarser tier counts point the same way, with every configuration beating the base model more often on low-resource than on high-resource languages in both regimes, rising for example from 33\% to 66\% for AdaMoE at 96 languages (Table~\ref{tab:rq4-resource}).

Because the base-model score appears on both axes, we also predict the gain on each benchmark from the base score on a \emph{different} benchmark, where the measurement noise is independent, and the relationship survives (Table~\ref{tab:rq4-cross-benchmark}). Nothing here is randomized, so we read all of this as a consistent association rather than a causal effect.

\section{Discussion and Conclusion}
\paragraph{What transfers is the region, not the ranking.}
The reduced-scale search gives a usable starting point. A low rank and a larger expert count are favoured on the ten-language data, and every configuration it promoted remained trainable at scale. What it does not give is an ordering, nor any guarantee of improving on the unadapted model, which places third of eight at 200 languages. The configuration ranked first at ten languages finished third in both broader regimes, and its ordering is uncorrelated with the one at 200. A short multilingual sweep is therefore a reasonable way to find viable configurations but an unreliable way to pick a winner. It also leaves open whether the configurations it discards are worse, since, of the two discarded configurations we retrained at scale, plain LoRA at $r{=}4$ placed second at 200 languages.

\paragraph{Method, configuration, and regime interact.}
MOELoRA and AdaMoE lead the controlled shared-rank comparison, yet different configurations lead once the selected factors are combined and the budget is extended, and different ones again in each broader regime. Combining selected values does not always help either, since the best mixed descendant fails to beat its preserved parent in two of six prespecified groups. One-factor trends therefore do not reliably compose, and no single leaderboard summarizes this method family. Routing does not cause this instability, as plain LoRA configurations also move between the two regimes. Practitioners should validate in the training regime they intend to deploy, because a shortlist assembled anywhere else does not rank correctly.

\paragraph{The effect is clear, but why it occurs is not.}
Across regimes, the effect of adaptation depends more on the benchmark than on the configuration. Belebele performance declines for all fourteen configuration--regime combinations. FLORES improves strongly at 96 training languages, although the benefit weakens at 200 languages for three of the seven configurations. Because plain LoRA behaves the same way on both benchmarks, the trade-off suggests a property of low-rank adaptation of a frozen multilingual model rather than of routing. \moelorafamily adaptation as configured here is therefore a targeted tool, not a general multilingual upgrade.

One explanation is that continued pretraining benefits the two benchmarks differently. FLORES measures how well the model fits reference translations, so the multilingual text modelling learned during CPT can transfer to it. In contrast, CPT does not directly train the in-context discrimination required by Belebele. Adaptation may therefore improve language-model fit while weakening behaviour already present in the base model. Such forgetting has been documented when pretrained models are extended to new languages~\cite{10.1609/aaai.v39i24.34805,li2025rethinking}, while low-rank adapters must balance learning new information against preserving existing capabilities~\cite{biderman2024lora}. Testing this would require a deeper look at sample-level performance, which we leave to future work. A short replay stage using data from the base model's original distribution as suggested by \citet{10.1609/aaai.v39i24.34805} might help recover original-language ability but still would not explain why adaptation trades comprehension for language-model fit.

\section*{Limitations}
Training all 166 configurations of the eight methods would cost more than 70{,}000 A100-hours across both broader scales. We therefore vary one design choice at a time and retrain only a shortlist, as is common practice~\cite{nllbteam2022languageleftbehindscaling,ustun-etal-2024-aya,zadouri2024pushing,gao-etal-2025-mola}. Apart from the plain-LoRA baselines, discarded configurations were never retrained at scale, so stronger configurations may lie outside the shortlist. Each configuration is also trained once, and close rankings may change on retraining. Runs are matched on wall clock rather than on tokens, so faster low-rank configurations also see more data (Table~\ref{tab:resource-accounting}). Benchmark coverage differs across stages and reaches only 136 of the 200 training languages. All results also come from a single base model and may not hold for others.

Future work can introduce seeds, change the base model, evaluate generated translations with metrics such as BLEU or COMET, or move selection closer to the target regime. We release the configurations, per-language results, and analysis code so that future work need not repeat the search.

\section*{Ethical Considerations}
This study involves no human subjects, annotation, or new data collection. The training text is derived from the web, and we apply no additional screening for personal or offensive content beyond the procedures used for the source corpora~\cite{NEURIPS2024_370df50c,penedo2025fineweb}. Residual personal information and harmful content may therefore remain, although we do not redistribute any training text. Appendix~\ref{sec:appendix-artifact} documents the licenses and intended uses of the artifacts used and released in this study. In total, the 146 training runs consumed approximately 5{,}150 A100-hours.

\section*{Acknowledgments}
This work has been funded by the Deutsche Forschungsgemeinschaft (DFG, German Research Foundation) – TRR 318/3 2026 – 438445824, by the Federal Ministry for Economic Affairs and Energy (BMWE) on the basis of a decision by the German Bundestag under the project ikDS (FKZ KK5175206LO4), and by the German Federal Ministry of Research, Technology and Space (BMFTR) within the project KI-Akademie OWL (16IS24057B) and Learn2RAG (01MK250104).

The authors gratefully acknowledge the computing time made available to them on the high-performance computer Noctua~2~\cite{noctua2} at the NHR Center Paderborn Center for Parallel Computing (PC2). This center is jointly supported by the Federal Ministry of Research, Technology and Space and the state governments participating in the National High-Performance Computing (NHR) joint funding program (\url{www.nhr-verein.de/en/our-partners}).

Generative AI tools (ChatGPT, Claude, Gemma, Qwen) were used for language refinement and code assistance. All scientific contributions and conclusions are solely those of the authors, who reviewed and take full responsibility for the final manuscript.

\bibliography{bib/references}

\clearpage
\appendix
\setcounter{table}{0}\renewcommand{\thetable}{A\arabic{table}}
\setcounter{figure}{0}\renewcommand{\thefigure}{A\arabic{figure}}

\setcounter{topnumber}{4}
\setcounter{dbltopnumber}{4}
\setcounter{totalnumber}{6}
\renewcommand{\topfraction}{0.95}
\renewcommand{\dbltopfraction}{0.95}
\renewcommand{\textfraction}{0.05}
\renewcommand{\bottomfraction}{0.75}
\renewcommand{\floatpagefraction}{0.80}
\renewcommand{\dblfloatpagefraction}{0.80}
\makeatletter
\setlength{\@fptop}{0pt}
\setlength{\@fpsep}{10pt plus 0fil}
\setlength{\@fpbot}{0pt plus 1fil}
\setlength{\@dblfptop}{0pt}
\setlength{\@dblfpsep}{10pt plus 0fil}
\setlength{\@dblfpbot}{0pt plus 1fil}
\makeatother

\section{Experimental Details}\label{sec:appendix}
In the tables below, $n$ is the number of paired evaluation cases behind a test, and stars mark the $p$-value in the same column, * for $p<.05$, ** for $p<.01$, *** for $p<.001$.

\subsection{Language Selection and Resource Labels}\label{sec:appendix-data}
Two released notebooks document how the training languages were chosen and how each language was labelled by resource level.

\paragraph{Resource labels.} Resource labels are generated by \artifactfile{data_sample/notebook/fineweb2_analysis_and_subsampling.ipynb}, which builds the language table used throughout the paper. It starts from the FineWeb2 language distribution, drops the subsets marked as removed, and joins an external resource-category list from \citet{chang-etal-2024-multilinguality}. Its original categories establish document-count boundaries, which are merged into the three levels we report, high, medium, and low, and applied to every retained FineWeb2 subset after sorting by document count.

\paragraph{Language selection.} The language selection is generated by \artifactfile{data_sample/notebook/lang2vec.ipynb}, which chooses the languages we train on. It reads the labelled table, keeps the languages with more than a thousand documents in FineWeb2, and intersects them with the languages for which lang2vec provides a genetic distance~\cite{littell-etal-2017-uriel}. From that pool it applies greedy farthest-point selection, repeatedly adding the language furthest from those already picked, so that each smaller set is contained in the next larger one. It also reports how much of Belebele and FLORES each set covers.

\subsection{Method and Grid Coverage}\label{sec:appendix-coverage}
\begin{table*}[tp]
\centering
\small
\setlength{\tabcolsep}{4pt}
\begin{tabular}{lcccl}
\toprule
Stage & Languages & Hours/run & Runs & Purpose \\
\midrule
One-axis rank sweep & 10 & 3.6 & 47 & Effect of rank \\
One-axis expert-count sweep & 10 & 3.6 & 63 & Effect of expert count \\
Mixed follow-up & 10 & 7.2 & 22 & Joint configurations and shortlist \\
Broader validation (96 languages) & 96 & 34.5 & 7 & Test the shortlist at scale \\
Broader validation (200 languages) & 200 & 72 & 7 & Test the shortlist at scale \\
\midrule
Total & -- & -- & 146 & 129 unique configurations; 10{,}842 evaluation runs \\
\bottomrule
\end{tabular}
\caption{Stages of the study, with training coverage, per-run budget, and run counts.}
\label{tab:study-funnel}
\end{table*}

\begin{table*}[tp]
\centering
\footnotesize
\setlength{\tabcolsep}{4pt}
\resizebox{\textwidth}{!}{%
\begin{tabular}{llcllc}
\toprule
Method & Rank search & Runs & Expert-count search ($E$:~top-$k$) & Runs & Panel \\
\midrule
AdaMoE & $r \in \{2, 4, 8, 16, 32, 64, 128\}$ at $E{=}8$, $k{=}6$ & 7 & $2$:~1; $4$:~1, 2; $8$:~1, 2, 4; $16$:~1, 2, 4, 8; $32$:~1, 2, 4, 8, 16 & 15 & yes \\
AdaMoLE & -- & -- & $2$, $4$, $8$, $16$ & 4 & no \\
HMoRA & $r \in \{2, 4, 8, 16, 32, 64, 128\}$ at $E{=}8$, $k{=}2$ & 7 & $2$:~1; $4$:~1, 2; $8$:~1, 2, 4; $16$:~1, 2, 4, 8; $32$:~1, 2, 4, 8, 16 & 15 & yes \\
HydraLoRA & $r \in \{2, 4, 8, 16, 32, 64, 128, 256\}$ & 8 & $2$, $4$, $8$, $16$, $32$ & 5 & yes \\
LoRA & $r \in \{2, 4, 8, 16, 32, 64, 128, 256, 512, 1024\}$ & 10 & -- & -- & yes \\
MOELoRA & $r \in \{2, 4, 8, 16, 32, 64\}$ at $E{=}8$, $k{=}2$ & 6 & $2$:~1; $4$:~1, 2; $8$:~1, 2, 4; $16$:~1, 2, 4, 8; $32$:~1, 2, 4, 8 & 14 & yes \\
MoV & no rank at $E{=}8$ & 1 & $2$, $4$, $8$, $16$, $32$ & 5 & no \\
MTL-LoRA & $r \in \{2, 4, 8, 16, 32, 64, 128, 256\}$ & 8 & $2$, $4$, $8$, $16$, $32$ & 5 & yes \\
\midrule
Total &  & 47 &  & 63 &  \\
\bottomrule
\end{tabular}
}
\caption{Every configuration trained in the two one-factor searches.}
\label{tab:config-grid}
\end{table*}

Table~\ref{tab:study-funnel} summarizes the five stages of the study and their run counts. Table~\ref{tab:config-grid} enumerates the grid itself, so that the 47 and 63 runs of the two one-factor searches can be reconstructed method by method. In it, $r$ is the rank of one expert, $E$ the number of experts, and $k$ the number of experts routed to per token, which methods without a router do not define. The rank search holds $E$ and $k$ fixed, and the expert-count search holds $r{=}2$. The Panel column marks the six methods with complete rank coverage, which are the ones compared in Table~\ref{tab:rq2-family}. The grid reports observed coverage and does not imply that every method was evaluated on the same grid. Method-specific configuration files, run identifiers, and raw benchmark rows accompany the released analysis.

\paragraph{What the searches leave out.} Four kinds of gap produce the uneven grids. First, some methods do not define a factor. Plain LoRA trains one adapter and so has no expert count, which is why it appears in the rank search only and why its two best ranks there, rather than a rank--expert pair, are what carries into the broader regimes; and \baseline{MoV} scales activation vectors rather than low-rank matrices and so has no rank, which is why it contributes a single run to the rank search. Second, some routing settings are degenerate. The expert-count search varies top-$k$ over the powers of two up to $E/2$, because $k{=}E$ routes to every expert and removes the routing being studied, and \baseline{MoV} is trained dense for the same reason. Third, some configurations did not fit the hardware. \baseline{AdaMoLE} at 32 experts exhausted memory on four A100 40GB GPUs at a per-device batch size of one, under both the all-target and the attention-only adapter placement, and \baseline{MOELoRA} at 32 experts with top-$k$ 16 was likewise infeasible, which is why those two cells are missing. Fourth, one method's grid did not train reliably. \baseline{AdaMoLE} was not run in the rank search, because its higher-rank configurations did not train stably, so it is excluded from the shared rank panel.

\paragraph{CoLA.} We also implemented and trained CoLA~\cite{zhou-etal-2025-cola}, with eight runs in the rank search and five in the expert-count search. Those runs failed early, and we excluded the method rather than report it on unequal terms. CoLA is therefore absent from every count, table, and test reported here, and the 132 search runs of Section~\ref{sec:stages} are the 145 trained search runs minus these 13. The released workbooks retain the CoLA rows.

Evaluation coverage is smaller than training coverage at every stage, so we state the two separately. The ten-language stages evaluate nine languages, giving 42 task and language cases per model. The broader regimes evaluate 80 languages (249 cases) after training on 96, and 136 languages (408 cases) after training on 200.

\subsection{Training Budget and Language Mixture}\label{sec:appendix-budget}
\begin{table*}[tp]
\centering
\small
\setlength{\tabcolsep}{4pt}
\begin{tabular}{lccrrrrrrrrr}
\toprule
Configuration & $E$ & $k$ & $r$ & $\alpha$ & Train.\ & Active & Steps & Tokens & Tok./s & Corpus & GPU \\
 & & & & & (M) & (M) & (k) & (B) & (k) & (\%) & (h) \\
\midrule
\multicolumn{12}{l}{\textit{96 training languages}} \\
MTL-LoRA r8/16 & 16 & dense & 8 & 16 & 188.0 & 186.1 & 20.0 & 0.98 & 8.0 & 63 & 136 \\
MOELoRA r2/16 k8 & 16 & 8 & 32 & 64 & 27.3 & 13.6 & 7.6 & 0.37 & 3.0 & 24 & 136 \\
HydraLoRA r2 & 8 & dense & 2 & 4 & 34.5 & 34.5 & 10.3 & 0.51 & 4.1 & 33 & 136 \\
MOELoRA r2/32 k8 & 32 & 8 & 64 & 128 & 54.5 & 13.6 & 6.7 & 0.33 & 2.7 & 21 & 136 \\
AdaMoE r2/8 k6 & 8 & 6 & 2 & 4 & 71.8 & 30--61 & 6.5 & 0.32 & 2.6 & 21 & 136 \\
LoRA r2 & 1 & -- & 2 & 4 & 5.2 & 5.2 & 24.1 & 1.19 & 9.7 & 77 & 136 \\
LoRA r4 & 1 & -- & 4 & 8 & 10.5 & 10.5 & 24.1 & 1.19 & 9.7 & 77 & 136 \\
\addlinespace
\multicolumn{12}{l}{\textit{200 training languages}} \\
MTL-LoRA r8/16 & 16 & dense & 8 & 16 & 190.0 & 186.1 & 30.6 & 1.50 & 5.8 & 38 & 287 \\
MOELoRA r2/16 k8 & 16 & 8 & 32 & 64 & 27.3 & 13.6 & 15.9 & 0.78 & 3.0 & 20 & 288 \\
HydraLoRA r2 & 8 & dense & 2 & 4 & 34.5 & 34.5 & 21.8 & 1.07 & 4.1 & 27 & 288 \\
MOELoRA r2/32 k8 & 32 & 8 & 64 & 128 & 54.5 & 13.6 & 10.2 & 0.50 & 1.9 & 13 & 288 \\
AdaMoE r2/8 k6 & 8 & 6 & 2 & 4 & 71.8 & 30--61 & 10.4 & 0.51 & 2.0 & 13 & 288 \\
LoRA r2 & 1 & -- & 2 & 4 & 5.2 & 5.2 & 48.9 & 2.40 & 9.3 & 61 & 287 \\
LoRA r4 & 1 & -- & 4 & 8 & 10.5 & 10.5 & 48.7 & 2.40 & 9.3 & 60 & 287 \\
\bottomrule
\end{tabular}
\caption{Training budget consumed by each configuration in the broader regimes.}
\label{tab:resource-accounting}
\end{table*}

\begin{table}[tbp]
\centering
\small
\setlength{\tabcolsep}{2pt}
\resizebox{\columnwidth}{!}{%
\begin{tabular}{lrrrrrrr}
\toprule
Training set & Labels & Ex.\ & Top-1 & Top-5 & 50\% & 90\% & Min \\
 & & (M) & (\%) & (\%) & & & (\%) \\
\midrule
10 (search) & 13 & 0.10 & 35.5 & 85.9 & 2 & 7 & 0.0040 \\
10 (follow-up) & 13 & 0.40 & 35.5 & 85.9 & 2 & 7 & 0.0040 \\
96 & 102 & 1.51 & 23.3 & 71.8 & 3 & 16 & 0.0001 \\
200 & 215 & 3.88 & 18.2 & 57.4 & 5 & 26 & 0.0001 \\
\bottomrule
\end{tabular}
}
\caption{Composition of each prepared training set.}
\label{tab:language-mixture}
\end{table}

Table~\ref{tab:resource-accounting} reports the training budget of every configuration in both broader regimes. $E$ is the expert count and $k$ the number of experts routed per token, or \emph{dense} where a method combines all of them. Steps counts optimizer steps, Tokens the input token positions processed, Tok./s the throughput, Corpus the share of the training set seen once, and GPU the hours on the run's four A100 40GB GPUs. Every run in a regime received the same wall clock, so the spread in steps and tokens is the throughput difference between methods. That spread is large. At 200 training languages the configurations process between 0.50 and 2.40 billion token positions, or between 13\% and 61\% of one pass over the training set. We report several indicators because they do not agree with each other~\cite{dehghani2022efficiency}. The configuration with the fewest trainable parameters sees the most data, and the configuration with the most trainable parameters is the fastest of the routed ones.

The same table gives the rank $r$ and scaling hyperparameter $\alpha$ of every configuration, as the implementation counts them. Across all 140 training runs with a low-rank update, $\alpha$ is twice that rank. \baseline{MOELoRA} counts the total rank budget, so its configurations carry $r=32$ and $r=64$ although the paper reports the per-expert rank of 2. Every implementation except \baseline{HMoRA} scales the update by $\alpha/r$, which is therefore $2$ in every run. For these methods the rank sweep does not change update magnitude. \baseline{HMoRA} specifically defines its own ratio of $\alpha$ to $r$, scaling the update by $\alpha/\sqrt{r}$, and we use that definition instead. With $\alpha=2r$ its factor is $2\sqrt{r}$, from 2.8 at $r=2$ to 22.6 at $r=128$, so its seven rank-sweep runs vary rank and update magnitude together. Its expert-count runs all use $r=2$ and share one factor. \baseline{HMoRA} is also not among the configurations carried to the broader regimes. \baseline{MoV} scales vectors rather than low-rank matrices and has no $\alpha$. The rank, $\alpha$, and applied scaling factor of every run are released with the run metadata.

The Train.\ and Active columns separate trainable adapter parameters from those reached on a forward pass, since an expert that is never selected for a token contributes to the first but not the second. The two \baseline{MOELoRA} configurations activate 13.6M of the 27.3M and 54.5M they train, because they route to 8 of 16 and 8 of 32 experts. \baseline{HydraLoRA} and \baseline{MTL-LoRA} combine every up-projection with dense weights, so their two counts nearly coincide, as they do exactly for plain LoRA. \baseline{AdaMoE} selects a variable number of experts per token through its null experts, so its count is given as a range rather than a point value. Counts are derived from the saved checkpoint shapes together with each run's routing settings, and they are parameter counts rather than FLOP estimates. The frozen backbone contributes a further 8.03B parameters to every configuration and is excluded from both columns.

Table~\ref{tab:language-mixture} gives the composition of each training set. Labels counts language--script labels, which exceed the nominal language count because a language can appear in more than one script, and Ex.\ counts packed training sequences of 1{,}024 tokens. Top-1 and Top-5 give the share of sequences held by the largest labels, and the 50\% and 90\% columns the number of labels needed to reach that share. Construction draws the same fraction of every language's available documents, 0.0025 at 96 training languages and 0.005 at 200, so the mixture reproduces FineWeb2's own distribution without upsampling low-resource languages. At 200 training languages five language--script labels account for half of the training sequences and 26 account for nine tenths, while the smallest label contributes about one sequence in a million. English is the one deliberate exception, set to the rounded median document target of the selected non-English entries rather than to its corpus share, which leaves it with 31 training documents at 200 languages. These are shares of the prepared split rather than measured per-run consumption, so they describe what each run could draw on and not what it saw. The broader-regime results hold for this mixture, which is one reason not to read them as the best achievable at 200 languages. Per-language document, sequence, and token counts are released with the analysis.

\subsection{Complete RQ1 Results}\label{sec:appendix-rq1}
\begin{table}[tbp]
\centering
\small
\begin{tabular}{lcc}
\toprule
Factor & Selection analysis & Follow-up values \\
\midrule
Rank $r$ & 2 (next: 8) & 2, 8 \\
Expert count & 32 (next: 16) & 32, 16 \\
\bottomrule
\end{tabular}
\caption{Design-factor values carried into the mixed follow-up.}
\label{tab:rq1-design}
\end{table}

\begin{table}[tbp]
\centering
\small
\setlength{\tabcolsep}{3pt}
\begin{tabular}{lccccc}
\toprule
Setting & Belebele & FLORES & SIB-200 & Taxi1500 & Overall \\
\midrule
\multicolumn{6}{l}{\textit{Rank $r$}} \\
2 & 2.76 & \textbf{2.00} & \textbf{2.89} & \textbf{2.31} & \textbf{2.49} \\
4 & 2.85 & 2.37 & 3.41 & 2.62 & 2.81 \\
8 & \textbf{2.41} & 2.36 & 3.32 & 2.79 & 2.72 \\
16 & 3.01 & 3.85 & 3.63 & 3.46 & 3.49 \\
32 & 4.94 & 4.83 & 3.60 & 4.60 & 4.49 \\
64 & 5.03 & 5.58 & 4.15 & 5.22 & 5.00 \\
\addlinespace
\multicolumn{6}{l}{\textit{Expert count}} \\
2 & 3.16 & 3.59 & 3.40 & 3.68 & 3.46 \\
4 & 2.80 & 2.89 & 2.67 & 3.00 & 2.84 \\
8 & 3.76 & 2.94 & 3.27 & 3.47 & 3.36 \\
16 & 2.81 & 3.50 & \textbf{2.19} & 2.51 & 2.75 \\
32 & \textbf{2.47} & \textbf{2.08} & 3.48 & \textbf{2.33} & \textbf{2.59} \\
\bottomrule
\end{tabular}
\caption{Average position of each rank and expert count on the balanced panels. Lower is better, best per column within each factor in bold.}
\label{tab:rq1-full-ranks}
\end{table}

\begin{table}[tbp]
\centering
\small
\setlength{\tabcolsep}{3pt}
\begin{tabular}{lrrrr}
\toprule
Benchmark & $n$ & $Q$ & $W$ & $p_{\mathrm{Holm}}$ \\
\midrule
\multicolumn{5}{l}{\textit{Rank $r$}} \\
Belebele & 54 & 106.24 & 0.393 & $7.7\!\times\!10^{-21}$*** \\
FLORES & 108 & 341.43 & 0.632 & $4.9\!\times\!10^{-71}$*** \\
SIB-200 & 54 & 13.60 & 0.050 & $0.018$* \\
Taxi1500 & 36 & 74.52 & 0.414 & $2.3\!\times\!10^{-14}$*** \\
\addlinespace
\multicolumn{5}{l}{\textit{Expert count}} \\
Belebele & 54 & 20.87 & 0.097 & $6.7\!\times\!10^{-4}$*** \\
FLORES & 108 & 62.99 & 0.146 & $2.7\!\times\!10^{-12}$*** \\
SIB-200 & 54 & 28.47 & 0.132 & $3.0\!\times\!10^{-5}$*** \\
Taxi1500 & 36 & 20.54 & 0.143 & $6.7\!\times\!10^{-4}$*** \\
\bottomrule
\end{tabular}
\caption{Friedman tests for the balanced rank and expert-count panels. $Q$ is the Friedman statistic and $W$ Kendall's coefficient of concordance.}
\label{tab:rq1-omnibus}
\end{table}

Table~\ref{tab:rq1-full-ranks} gives the average position of every rank and every expert count on the balanced panels, blocking by method so that each method contributes its own ordering. Table~\ref{tab:rq1-design} states the two values of each factor that this analysis carried into the mixed follow-up, and Table~\ref{tab:rq1-omnibus} reports the corresponding Friedman tests, with Holm correction applied over the four benchmarks separately for each factor. These tests are exploratory, because the same observations were used to select the values carried to the follow-up.

\subsection{Complete RQ2 Results}\label{sec:appendix-rq2}
\begin{table*}[tp]
\centering
\small
\begin{tabular}{lcccrrrrr}
\toprule
Method & Rank $r$ & Experts & Top-$k$ & Belebele & FLORES & SIB-200 & Taxi1500 & Overall \\
\midrule
MTL-LoRA & 8 & 16 & -- & 7.11 & 10.28 & 4.33 & 7.42 & \textbf{7.28} \\
MOELoRA & 2 & 16 & 8 & 6.44 & 4.94 & 11.61 & 6.92 & 7.48 \\
HydraLoRA & 2 & -- & -- & 10.83 & 9.89 & \textbf{4.11} & 5.17 & 7.50 \\
MOELoRA & 2 & 32 & 8 & \textbf{4.17} & 9.22 & 14.50 & \textbf{4.83} & 8.18 \\
AdaMoE & 2 & 8 & 6 & 10.83 & \textbf{4.33} & 8.94 & 12.83 & 9.24 \\
\bottomrule
\end{tabular}
\caption{The five configurations carried to the broader validation, with their average positions among all 22 follow-up configurations. Lower is better, best per column in bold.}
\label{tab:rq2-followup}
\end{table*}

\begin{table*}[tp]
\centering
\small
\setlength{\tabcolsep}{3pt}
\begin{tabular}{rlllcccrrrr}
\toprule
\# & Run ID & Method & Rank $r$ & Experts & Top-$k$ & Belebele & FLORES & SIB-200 & Taxi1500 & Overall \\
\midrule
\textbf{1} & \textbf{\texttt{mtllora\_r8\_nup16}} & \textbf{MTL-LoRA} & \textbf{8} & \textbf{16} & \textbf{--} & \textbf{7.11} & \textbf{10.28} & \textbf{4.33} & \textbf{7.42} & \textbf{7.28} \\
\textbf{2} & \textbf{\texttt{moelora\_tr32\_e16\_k8}} & \textbf{MOELoRA} & \textbf{2} & \textbf{16} & \textbf{8} & \textbf{6.44} & \textbf{4.94} & \textbf{11.61} & \textbf{6.92} & \textbf{7.48} \\
\textbf{3} & \textbf{\texttt{hydralora\_r2}} & \textbf{HydraLoRA} & \textbf{2} & \textbf{--} & \textbf{--} & \textbf{10.83} & \textbf{9.89} & \textbf{4.11} & \textbf{5.17} & \textbf{7.50} \\
\textbf{4} & \textbf{\texttt{moelora\_tr64\_e32\_k8}} & \textbf{MOELoRA} & \textbf{2} & \textbf{32} & \textbf{8} & \textbf{4.17} & \textbf{9.22} & \textbf{14.50} & \textbf{4.83} & \textbf{8.18} \\
\textbf{5} & \textbf{\texttt{mola\_tr16\_e8\_k6}} & \textbf{AdaMoE} & \textbf{2} & \textbf{8} & \textbf{6} & \textbf{10.83} & \textbf{4.33} & \textbf{8.94} & \textbf{12.83} & \textbf{9.24} \\
6 & \texttt{moelora\_tr64\_e32\_k4} & MOELoRA & 2 & 32 & 4 & 9.00 & 8.33 & 11.28 & 8.92 & 9.38 \\
7 & \texttt{moelora\_tr256\_e32\_k8} & MOELoRA & 8 & 32 & 8 & 7.06 & 4.06 & 16.00 & 13.58 & 10.17 \\
8 & \texttt{mtllora\_r8} & MTL-LoRA & 8 & -- & -- & 8.17 & 19.83 & 6.22 & 8.58 & 10.70 \\
9 & \texttt{moelora\_tr128\_e16\_k8} & MOELoRA & 8 & 16 & 8 & 8.22 & 9.56 & 14.11 & 14.08 & 11.49 \\
10 & \texttt{mtllora\_r8\_nup32} & MTL-LoRA & 8 & 32 & -- & 12.00 & 12.00 & 8.44 & 13.58 & 11.51 \\
11 & \texttt{mola\_tr64\_e32\_k4} & AdaMoE & 2 & 32 & 4 & 15.00 & 9.56 & 14.89 & 7.08 & 11.63 \\
12 & \texttt{hydralora\_r2\_heads32} & HydraLoRA & 2 & 32 & -- & 9.61 & 11.50 & 9.50 & 16.25 & 11.72 \\
13 & \texttt{hydralora\_r2\_heads16} & HydraLoRA & 2 & 16 & -- & 16.83 & 10.67 & 7.94 & 11.67 & 11.78 \\
14 & \texttt{mola\_tr256\_e32\_k4} & AdaMoE & 8 & 32 & 4 & 17.56 & 12.39 & 11.06 & 7.08 & 12.02 \\
15 & \texttt{moelora\_tr256\_e32\_k4} & MOELoRA & 8 & 32 & 4 & 9.11 & 12.56 & 13.83 & 13.75 & 12.31 \\
16 & \texttt{moelora\_tr64\_e8\_k2} & MOELoRA & 8 & 8 & 2 & 18.44 & 6.28 & 21.67 & 3.50 & 12.47 \\
17 & \texttt{mola\_tr128\_e16\_k1} & AdaMoE & 8 & 16 & 1 & 13.44 & 20.94 & 4.44 & 12.67 & 12.87 \\
18 & \texttt{mola\_tr32\_e16\_k1} & AdaMoE & 2 & 16 & 1 & 13.72 & 17.22 & 10.72 & 12.17 & 13.46 \\
19 & \texttt{moelora\_tr16\_e8\_k2} & MOELoRA & 2 & 8 & 2 & 7.89 & 14.39 & 21.33 & 11.58 & 13.80 \\
20 & \texttt{mola\_tr32\_e16\_k8} & AdaMoE & 2 & 16 & 8 & 15.78 & 12.78 & 7.28 & 20.00 & 13.96 \\
21 & \texttt{mola\_tr128\_e16\_k8} & AdaMoE & 8 & 16 & 8 & 16.00 & 15.44 & 12.22 & 20.67 & 16.08 \\
22 & \texttt{mola\_tr64\_e8\_k6} & AdaMoE & 8 & 8 & 6 & 15.78 & 16.83 & 18.56 & 20.67 & 17.96 \\
\bottomrule
\end{tabular}
\caption{Average position of every mixed follow-up configuration. Lower is better, and bold marks the five configurations carried to the broader validation.}
\label{tab:rq2-followup-full}
\end{table*}

The 22 follow-up configurations are built as follows. We take the three best concrete configurations at each of the four selected factor values, then pair each method's selected rank with every expert count and routing setting that the same method contributed to the expert-count selection, and remove duplicates. Table~\ref{tab:rq2-followup-full} orders all 22 of them, and Table~\ref{tab:rq2-followup} repeats the five that were promoted. Table~\ref{tab:rq2-tests} reports the omnibus tests over all 22 configurations together with two-sided Wilcoxon tests among the promoted top three, with Holm correction applied over the four omnibus tests and, separately, over all twelve post-hoc tests. That family was fixed after the ranking was known, so it is less conservative than the all-pairs correction used in the broader regimes. Table~\ref{tab:rq2-mixing} isolates the effect of combining a selected rank with a selected expert count. It selects the best observed child within each prespecified parent group and is therefore descriptive. It should not be read as an unbiased estimate of the benefit of mixing design choices.

\subsection{Complete Broader-Regime Tests}\label{sec:appendix-rq3}
Table~\ref{tab:rq3-vs-baseline-tests} reports the seven comparisons against the base model for each benchmark within each broader regime, and Table~\ref{tab:rq3-omnibus} the corresponding omnibus tests, with Holm correction over the four benchmarks within each regime. For the pairwise tests, Holm correction covers all twenty-eight system pairs inside each regime and benchmark, not only the pairs shown here, and the remaining pairs are released with the analysis code. Correcting over every pair rather than over a shortlist keeps the size of the correction family independent of which systems come out on top, which makes these comparisons more conservative than a pre-planned finalist-versus-base design would have been.

Table~\ref{tab:rq4-resource} reports how often each configuration beats the base model, split by the resource level of the language, which follows the assignment described in Appendix~\ref{sec:appendix-data}. Low-resource languages dominate the evaluation population. Table~\ref{tab:rq4-cross-benchmark} is the independent-noise check for Figure~\ref{fig:rq4-gain-vs-base}. The base-model score appears on both axes of that figure, so a negative relationship could in principle be produced by measurement noise alone. Taking the base score from a different benchmark than the gain removes that path, and the negative relationship holds everywhere except Taxi1500.

\subsection{Released Code and Notebooks}\label{sec:appendix-artifact}
The complete artifact is available at \artifactlink. It contains the training and evaluation pipeline behind all 146 training runs and 10{,}842 evaluation runs, together with the per-method configuration files and run identifiers referred to throughout this appendix. A \nolinkurl{paper_artifact/} directory holds the analysis layer, which comprises the result workbooks, the scripts that turn them into every table and figure of this paper, and the per-question notebooks listed below under \nolinkurl{paper_artifact/notebooks/}.

\paragraph{Configuration names and run identifiers.} The paper names a configuration by its method, per-expert rank, expert count, and top-$k$, while the released runs keep the identifiers the pipeline generated, which encode the \emph{total} rank budget and carry two historical prefixes. Table~\ref{tab:naming} maps between them for the configurations carried to the broader regimes. In an identifier, \texttt{tr} is the total rank budget, which the paper divides by the expert count \texttt{e} to give the per-expert rank $r$, so \texttt{tr32\_e16} is $r{=}2$ with $E{=}16$. \texttt{k} is the top-$k$, and \texttt{nup} is the number of up-projections that MTL-LoRA calls an expert count. The \texttt{mola} prefix marks the AdaMoE implementation, whose runs appear as \texttt{adamoe\_1\_*} in the expert-count sweep. The same rule applies to every identifier in Table~\ref{tab:rq2-followup-full}.

\paragraph{Per-question notebooks.} Each notebook reads the released workbooks and reproduces, in reporting order, the tables and figures for one research question.
\begin{itemize}[leftmargin=1.2em,itemsep=2pt,parsep=0pt,topsep=3pt]
\item \artifactnb{00_study_overview.ipynb} covers the study funnel, method coverage, and configuration grid (Appendix~\ref{sec:appendix-coverage}).
\item \artifactnb{01_rq1_design_factors.ipynb} covers rank and expert count (Section~\ref{sec:rq1}, Appendix~\ref{sec:appendix-rq1}).
\item \artifactnb{02_rq2_method_comparison.ipynb} covers the shared-panel method comparison, the 22 follow-up configurations, and their statistical tests (Section~\ref{sec:rq2}, Appendix~\ref{sec:appendix-rq2}).
\item \artifactnb{03_rq3_broader_regimes.ipynb} covers the 96- and 200-language regimes, the selection-transfer comparison, and the complete pairwise tests summarized in Appendix~\ref{sec:appendix-rq3} (Section~\ref{sec:rq3}).
\item \artifactnb{04_rq4_localization.ipynb} covers gains and regressions by task, translation direction, and resource level (Section~\ref{sec:rq4}).
\end{itemize}

\paragraph{Licensing and intended use.} Every artifact we build on is used for non-commercial research, which is the use its terms provide for. The base model is used under the Llama~3.1 Community License, the training text under the Open Data Commons Attribution License that covers FineWeb and FineWeb2, and Belebele, FLORES-200, and SIB-200 under CC~BY-SA~4.0. Taxi1500 is used under the terms its authors distribute it with, which restrict it to research use. The evaluation harness is used under its MIT license. We redistribute none of this material. What we release is our own configurations, run records, per-language scores, and analysis code, under CC~BY~4.0 for the results and the MIT license for the code, so that the terms above continue to govern the underlying artifacts.

\paragraph{Reproducing the reported numbers.}\leavevmode\nolinkurl{paper_artifact/scripts/run_all.py} regenerates every table and figure from the workbooks in \nolinkurl{paper_artifact/input/}, writing the \LaTeX{} sources and their underlying CSV files to \nolinkurl{paper_artifact/output/}. The pinned environment is given in \nolinkurl{paper_artifact/env/environment.yml}. The workbooks retain the rows excluded from the analysis, including the CoLA runs and the unbalanced grid cells, so the filtering decisions described above can be inspected rather than taken on trust.

\begin{table*}[tp]
\centering
\footnotesize
\setlength{\tabcolsep}{3pt}
\begin{tabular}{lcllrrrrr}
\toprule
Method & Rank $r$ & Preserved parent & Best mixed descendant & Belebele & FLORES & SIB-200 & Taxi1500 & Overall \\
\midrule
MOELoRA & 2 & \texttt{moelora\_tr16\_e8\_k2} & \texttt{moelora\_tr32\_e16\_k8} & +1.44 & +9.44 & +9.72 & +4.67 & \textbf{+6.32} \\
MOELoRA & 8 & \texttt{moelora\_tr64\_e8\_k2} & \texttt{moelora\_tr256\_e32\_k8} & +11.39 & +2.22 & +5.67 & -10.08 & \textbf{+2.30} \\
AdaMoE & 2 & \texttt{mola\_tr16\_e8\_k6} & \texttt{mola\_tr64\_e32\_k4} & -4.17 & -5.22 & -5.94 & +5.75 & -2.40 \\
AdaMoE & 8 & \texttt{mola\_tr64\_e8\_k6} & \texttt{mola\_tr256\_e32\_k4} & -1.78 & +4.44 & +7.50 & +13.58 & \textbf{+5.94} \\
HydraLoRA & 2 & \texttt{hydralora\_r2} & \texttt{hydralora\_r2\_heads32} & +1.22 & -1.61 & -5.39 & -11.08 & -4.22 \\
MTL-LoRA & 8 & \texttt{mtllora\_r8} & \texttt{mtllora\_r8\_nup16} & +1.06 & +9.56 & +1.89 & +1.17 & \textbf{+3.42} \\
\bottomrule
\end{tabular}
\caption{Effect of combining the selected rank and expert-count values. Cells are the parent's average position minus the descendant's, so positive values favour mixing and bold marks a positive overall effect.}
\label{tab:rq2-mixing}
\end{table*}

\begin{table*}[tp]
\centering
\small
\begin{tabular}{lrrrccc}
\toprule
Benchmark & $n$ & Kendall's $W$ & Omnibus $p_{\mathrm{Holm}}$ & \shortstack{MTL-LoRA\\--MOELoRA} & \shortstack{MTL-LoRA\\--HydraLoRA} & \shortstack{MOELoRA\\--HydraLoRA} \\
\midrule
Belebele & 9 & 0.410 & $4.2\!\times\!10^{-8}$*** & $\rightarrow$ $1$ & $\leftarrow$ $1$ & $\leftarrow$ $0.781$ \\
FLORES & 18 & 0.511 & $3.4\!\times\!10^{-29}$*** & $\rightarrow$ $0.011$* & $\leftarrow$ $1$ & $\leftarrow$ $0.002$** \\
SIB-200 & 9 & 0.612 & $1.3\!\times\!10^{-14}$*** & $\leftarrow$ $0.176$ & $=$ $1$ & $\rightarrow$ $0.078$ \\
Taxi1500 & 6 & 0.604 & $4.2\!\times\!10^{-8}$*** & $\leftarrow$ $1$ & $\rightarrow$ $1$ & $\rightarrow$ $1$ \\
\bottomrule
\end{tabular}
\caption{Omnibus and pairwise tests for the mixed follow-up. The omnibus test covers all 22 configurations, while the pairwise columns cover the top three of the five promoted, which is every pairing of those three. Arrows point toward the descriptively better configuration.}
\label{tab:rq2-tests}
\end{table*}

\begin{table*}[tp]
\centering
\small
\setlength{\tabcolsep}{3pt}
\begin{tabular}{lrrrrrr}
\toprule
 & \multicolumn{3}{c}{96 training languages} & \multicolumn{3}{c}{200 training languages} \\
\cmidrule(lr){2-4}\cmidrule(lr){5-7}
Configuration & Median diff. & Effect & $p_{\mathrm{Holm}}$ & Median diff. & Effect & $p_{\mathrm{Holm}}$ \\
\midrule
\multicolumn{7}{l}{\textit{Belebele}} \\
MTL-LoRA r8/16 & -0.0250 & -0.87 & $5.0\!\times\!10^{-5}$*** & -0.0222 & -0.90 & $9.9\!\times\!10^{-9}$*** \\
MOELoRA r2/16 k8 & -0.0200 & -0.85 & $6.3\!\times\!10^{-6}$*** & -0.0244 & -0.98 & $2.2\!\times\!10^{-10}$*** \\
HydraLoRA r2 & -0.0167 & -0.82 & $2.0\!\times\!10^{-4}$*** & -0.0233 & -0.92 & $3.1\!\times\!10^{-9}$*** \\
MOELoRA r2/32 k8 & -0.0228 & -0.82 & $2.2\!\times\!10^{-4}$*** & -0.0178 & -0.86 & $3.7\!\times\!10^{-8}$*** \\
AdaMoE r2/8 k6 & -0.0344 & -0.95 & $2.7\!\times\!10^{-8}$*** & -0.0344 & -0.99 & $4.0\!\times\!10^{-10}$*** \\
LoRA r2 & -0.0144 & -0.78 & $8.0\!\times\!10^{-5}$*** & -0.0178 & -0.89 & $2.1\!\times\!10^{-8}$*** \\
LoRA r4 & -0.0228 & -0.92 & $1.6\!\times\!10^{-7}$*** & -0.0267 & -0.91 & $3.9\!\times\!10^{-9}$*** \\
\addlinespace
\multicolumn{7}{l}{\textit{FLORES}} \\
MTL-LoRA r8/16 & +0.0058 & +0.78 & $2.5\!\times\!10^{-10}$*** & +0.0023 & +0.25 & $0.027$* \\
MOELoRA r2/16 k8 & +0.0059 & +0.94 & $1.1\!\times\!10^{-14}$*** & +0.0006 & +0.17 & $0.18$ \\
HydraLoRA r2 & +0.0047 & +0.73 & $5.9\!\times\!10^{-9}$*** & +0.0064 & +0.81 & $4.8\!\times\!10^{-18}$*** \\
MOELoRA r2/32 k8 & +0.0064 & +0.98 & $3.5\!\times\!10^{-16}$*** & +0.0019 & +0.40 & $8.9\!\times\!10^{-5}$*** \\
AdaMoE r2/8 k6 & +0.0056 & +0.79 & $1.3\!\times\!10^{-10}$*** & +0.0060 & +0.72 & $1.5\!\times\!10^{-14}$*** \\
LoRA r2 & +0.0052 & +0.87 & $1.2\!\times\!10^{-12}$*** & +0.0060 & +0.83 & $6.8\!\times\!10^{-19}$*** \\
LoRA r4 & +0.0058 & +0.84 & $7.6\!\times\!10^{-12}$*** & +0.0049 & +0.67 & $9.7\!\times\!10^{-13}$*** \\
\addlinespace
\multicolumn{7}{l}{\textit{SIB-200}} \\
MTL-LoRA r8/16 & -0.0441 & -0.60 & $0.003$** & -0.0000 & +0.06 & $1$ \\
MOELoRA r2/16 k8 & -0.0000 & -0.09 & $1$ & -0.0588 & -0.87 & $1.2\!\times\!10^{-10}$*** \\
HydraLoRA r2 & +0.0049 & +0.01 & $1$ & +0.0392 & +0.70 & $4.1\!\times\!10^{-7}$*** \\
MOELoRA r2/32 k8 & +0.0343 & +0.75 & $9.1\!\times\!10^{-5}$*** & -0.0147 & -0.10 & $1$ \\
AdaMoE r2/8 k6 & -0.0294 & -0.52 & $0.011$* & -0.0049 & -0.07 & $1$ \\
LoRA r2 & -0.0147 & -0.09 & $1$ & -0.1324 & -0.99 & $6.9\!\times\!10^{-14}$*** \\
LoRA r4 & -0.0490 & -0.78 & $2.0\!\times\!10^{-5}$*** & -0.0441 & -0.64 & $3.5\!\times\!10^{-6}$*** \\
\addlinespace
\multicolumn{7}{l}{\textit{Taxi1500}} \\
MTL-LoRA r8/16 & +0.0180 & +0.56 & $0.006$** & +0.0090 & +0.32 & $0.092$ \\
MOELoRA r2/16 k8 & +0.0270 & +0.90 & $4.6\!\times\!10^{-7}$*** & +0.0090 & +0.36 & $0.066$ \\
HydraLoRA r2 & -0.0000 & -0.36 & $0.288$ & +0.0090 & +0.16 & $1$ \\
MOELoRA r2/32 k8 & -0.0000 & +0.20 & $1$ & +0.0090 & +0.28 & $0.192$ \\
AdaMoE r2/8 k6 & +0.0090 & +0.23 & $1$ & -0.0000 & -0.10 & $1$ \\
LoRA r2 & +0.0270 & +0.72 & $4.2\!\times\!10^{-5}$*** & +0.0180 & +0.63 & $1.2\!\times\!10^{-5}$*** \\
LoRA r4 & -0.0000 & -0.08 & $1$ & +0.0721 & +1.00 & $2.2\!\times\!10^{-15}$*** \\
\bottomrule
\end{tabular}
\caption{Every configuration against the base model in the broader regimes. Median difference and rank-biserial effect are signed so that positive favours the configuration. Median differences are in each benchmark's own units and do not compare across benchmarks.}
\label{tab:rq3-vs-baseline-tests}
\end{table*}

\begin{table}[tbp]
\centering
\small
\setlength{\tabcolsep}{3pt}
\begin{tabular}{lrrrr}
\toprule
Benchmark & $n$ & $Q$ & $W$ & $p_{\mathrm{Holm}}$ \\
\midrule
\multicolumn{5}{l}{\textit{96 training languages}} \\
Belebele & 40 & 79.70 & 0.285 & $1.6\!\times\!10^{-14}$*** \\
FLORES & 100 & 142.71 & 0.204 & $4.1\!\times\!10^{-27}$*** \\
SIB-200 & 51 & 170.87 & 0.479 & $6.6\!\times\!10^{-33}$*** \\
Taxi1500 & 58 & 118.12 & 0.291 & $3.8\!\times\!10^{-22}$*** \\
\addlinespace
\multicolumn{5}{l}{\textit{200 training languages}} \\
Belebele & 65 & 148.73 & 0.327 & $7.5\!\times\!10^{-29}$*** \\
FLORES & 166 & 350.68 & 0.302 & $2.6\!\times\!10^{-71}$*** \\
SIB-200 & 84 & 363.88 & 0.619 & $5.3\!\times\!10^{-74}$*** \\
Taxi1500 & 93 & 225.34 & 0.346 & $9.7\!\times\!10^{-45}$*** \\
\bottomrule
\end{tabular}
\caption{Friedman tests over the eight systems in each broader regime. $Q$ is the Friedman statistic and $W$ Kendall's coefficient of concordance. Holm correction is applied over the four benchmarks within each regime.}
\label{tab:rq3-omnibus}
\end{table}

\begin{table}[tbp]
\centering
\small
\begin{tabular}{lccc}
\toprule
Configuration & High & Medium & Low \\
\midrule
\multicolumn{4}{l}{\textit{96 training languages} (17/8/55 languages)} \\
MTL-LoRA r8/16 & 40 & 40 & 70 \\
MOELoRA r2/16 k8 & 49 & 60 & 77 \\
HydraLoRA r2 & 32 & 51 & 62 \\
MOELoRA r2/32 k8 & 58 & 66 & 68 \\
AdaMoE r2/8 k6 & 33 & 40 & 66 \\
LoRA r2 & 44 & 49 & 78 \\
LoRA r4 & 33 & 40 & 61 \\
\addlinespace
\multicolumn{4}{l}{\textit{200 training languages} (23/13/100 languages)} \\
MTL-LoRA r8/16 & 41 & 42 & 58 \\
MOELoRA r2/16 k8 & 35 & 41 & 41 \\
HydraLoRA r2 & 51 & 54 & 69 \\
MOELoRA r2/32 k8 & 36 & 27 & 61 \\
AdaMoE r2/8 k6 & 39 & 44 & 59 \\
LoRA r2 & 46 & 47 & 55 \\
LoRA r4 & 45 & 51 & 68 \\
\bottomrule
\end{tabular}
\caption{Percentage of evaluation cases won against the base model, by the resource level of the language.}
\label{tab:rq4-resource}
\end{table}

\begin{table}[tbp]
\centering
\small
\setlength{\tabcolsep}{3pt}
\begin{tabular}{lrrrr}
\toprule
Gain on & \multicolumn{4}{c}{Base-model score taken from} \\
\cmidrule(lr){2-5}
 & Belebele & FLORES & SIB-200 & Taxi1500 \\
\midrule
\multicolumn{5}{l}{\textit{96 training languages}} \\
Belebele & -- & -0.52*** & -0.53*** & -0.67*** \\
FLORES & -0.58*** & -- & -0.48*** & -0.56** \\
SIB-200 & -0.69*** & -0.76*** & -- & -0.50** \\
Taxi1500 & +0.01 & -0.03 & +0.10 & -- \\
\addlinespace
\multicolumn{5}{l}{\textit{200 training languages}} \\
Belebele & -- & -0.56*** & -0.58*** & -0.61*** \\
FLORES & -0.46*** & -- & -0.29** & -0.41** \\
SIB-200 & -0.67*** & -0.69*** & -- & -0.47** \\
Taxi1500 & -0.18 & -0.25 & -0.04 & -- \\
\bottomrule
\end{tabular}
\caption{Spearman correlation between the gain on one benchmark and the base-model score on another. The diagonal is empty because it would take both values from the same benchmark.}
\label{tab:rq4-cross-benchmark}
\end{table}

\begin{table}[tbp]
\centering
\small
\setlength{\tabcolsep}{4pt}
\begin{tabular}{ll}
\toprule
Name in this paper & Released run identifier \\
\midrule
MTL-LoRA r8/16 & \texttt{mtllora\_r8\_nup16} \\
MOELoRA r2/16 k8 & \texttt{moelora\_tr32\_e16\_k8} \\
HydraLoRA r2 & \texttt{hydralora\_r2} \\
MOELoRA r2/32 k8 & \texttt{moelora\_tr64\_e32\_k8} \\
AdaMoE r2/8 k6 & \texttt{mola\_tr16\_e8\_k6} \\
LoRA r2 & \texttt{lora\_r2} \\
LoRA r4 & \texttt{lora\_r4} \\
Base model & \texttt{baseline::l-3-1-8b} \\
\bottomrule
\end{tabular}
\caption{Configuration names in this paper and the released run identifiers.}
\label{tab:naming}
\end{table}

\end{document}